\documentclass[letterpaper, 10 pt, conference]{ieeeconf}  

\IEEEoverridecommandlockouts                              

\usepackage{graphics} 
\usepackage{hyperref}
\usepackage{booktabs}
\usepackage[dvipsnames]{xcolor}
\usepackage{xcolor}
\usepackage{xspace}
\usepackage{graphicx}
\usepackage{adjustbox}
\usepackage{amsmath}
\usepackage{amssymb}
\usepackage{cleveref}
\usepackage{caption}
\usepackage{subcaption}
\usepackage{algorithm}
\usepackage{algorithmic}
\usepackage{xcolor}
\usepackage{cuted}
\usepackage{circledsteps}
\usepackage[normalem]{ulem}

\usepackage{booktabs}        
\usepackage{graphicx}        
\usepackage[table]{xcolor}   
\usepackage{multirow}        

\newcommand{\ours}{PixelLoop\xspace}
\newcommand{\mnav}{MASt3R-Nav\xspace}
\newcommand{\master}{MASt3R\xspace}
\newcommand{\objreact}{ObjectReact\xspace}

\title{\LARGE \bf \ours: Shortcut Topological Navigation with Pixel-Level Loops}

\author{
Sarthak Chittawar$^{1 \dagger}$,
Vansh Garg$^{1}$,
Aditya Vadali$^{1}$,
Krish Pandya$^{1}$,\\[0.8ex]
Rohit Jayanti$^{1}$,
Sourav Garg$^{1}$ and
Madhava Krishna$^{1}$%
  \thanks{The authors acknowledge the support of Ati Motors for this project.}
\thanks{${\dagger}$ Corresponding Author.}%
  \thanks{
    $^{1}$ Robotics Research Center, IIIT-Hyderabad, India
  }%
}

\begin{document}

\maketitle

\thispagestyle{empty}
\pagestyle{empty}

\setlength{\stripsep}{0pt}
\preCutedStrip={\vspace{-1cm}}

\begin{strip}
    \centering
    \includegraphics[width=0.875\linewidth]{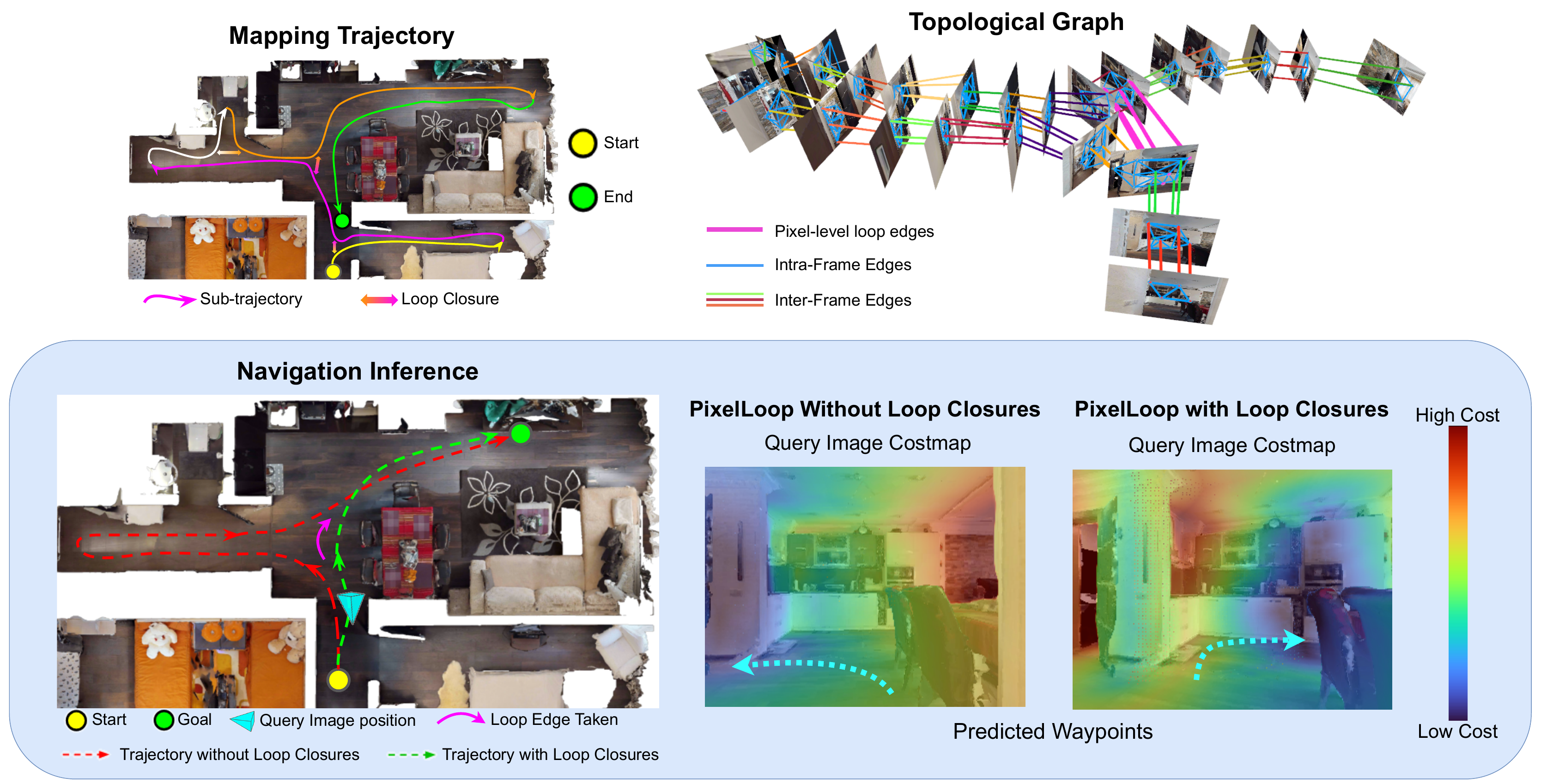}

    \captionof{figure}{\textbf{Top:} Start-to-goal navigation over large-scale mapping runs composed of multiple teach-and-repeat sub-trajectories (shown in different colours) is challenging without loop closures. \ours detects revisited regions and establishes pixel-level loop edges, integrating them into a densely loop-connected (shown as \textcolor{magenta}{magenta} edges) topological graph.
    \textbf{Bottom:} The resulting costmaps enable direct start-to-goal paths (\textcolor{green}{green}) via loop closures, whereas the no-loop configuration (\textcolor{red}{red}) produces longer detours. \textit{(Absolute poses are used for visualization only; our method operates entirely in relative 3D space.)}}
    \label{fig:teaser}
    \vspace{0.35cm}
\end{strip}

\preCutedStrip={}

\begin{abstract}
Although topological mapping and navigation have been studied extensively, the specific role and downstream effect of loop closures in purely topological representations has received relatively little attention. Importantly, loop closure over topological maps is distinct from loop closure over globally referenced trajectories and metric maps. Building on recent denser topologies grounded in pixel-level, relative 3D geometry, we propose \textbf{\ours} which introduces loop closures directly in pixel space. Unlike sparse image-level edges or pose-graph corrections in SLAM, our pixel-level closures act as dense topological shortcuts that alter planning connectivity and cost propagation rather than merely aligning coordinates. This dense connectivity enables stable any-point-to-any-point navigation and produces costmaps that align accurately with geometric shortest paths. In particular, we showcase the distinct advantage of applying loop closures to fine-grained pixel topologies rather than image-level topologies. Across extensive simulated experiments, \ours achieves over 35\% absolute improvement in both Success Rate and SPL compared to image-relative baselines, with the largest gains in scenarios requiring shortcut exploitation. Results are further validated through real-world mobile robot deployments, demonstrating that dense pixel-level loop closures provide a practical and robust foundation for topological visual navigation. \newline
Project Page: https://pixelloop-nav.github.io/
\end{abstract}

\section{INTRODUCTION}
Visual navigation has recently advanced through image-conditioned policies and trajectory-centric topological representations that bypass explicit, globally consistent 3D maps~\cite{garg2025objectreact, shah2023gnm, garg2024robohop, sridhar2024nomad}. A more recent method, \mnav~\cite{anonymous2026mast3rnav}, has demonstrated that pixel-level geometric correspondences can be leveraged to construct topological graphs grounded in 3D relative mapping, enabling robust path planning and control. However, topological navigation is typically constrained to \emph{teach-and-repeat} settings through a one-dimensional representation of a route as an image sequence, restricting navigation to replaying or locally adapting a single trajectory. In the absence of globally grounded reference frames, relative distances between two topologies become constrained or biased based on the order in which they appear in the reference run. These sequential biases carry forward to subsequent inference runs, making trajectories to goal queries unnaturally long with increasing failure rates.

The core limitation lies in topological connectivity. Topology induced maps, such as a sequential collection of images (without a grounded pose reference), encode sequential adjacency but lack mechanisms to structurally link regions that are geometrically proximate but temporally distant. Consequently, naively planning paths along such a sequence forces the agent to retrace convoluted paths rather than taking obvious physical shortcuts. As illustrated in Fig.~\ref{fig:teaser}, multiple teach-and-repeat sub-trajectory runs (shown in different colours) remain disconnected without loop closures. Two rooms separated by a single wall (or common space) might be adjacent in 3D space but hundreds of frame-nodes apart in a purely sequential graph, forcing a severe detour (red trajectory). The fundamental problem is that the geometry of the planning graph reflects historical trajectory rather than true shortest-path structure.

Furthermore, the role of loop closures in topological representations remains largely unexplored, where early image-based probabilistic topological approaches such as FAB-MAP~\cite{FABMAP2010fab} have mainly focused on mapping rather than path planning or navigation. Unlike globally-referenced metric SLAM systems, loop closures in topological maps strongly affect the planning structure. In this work, we study this phenomenon in the context of fine-grained pixel-level topology and demonstrate that loop closures applied at the pixel level provide substantially stronger navigation performance than closures applied over image-level topologies.

We propose \textbf{\ours}, an arbitrary start-to-goal navigation stack built upon 3D relative mapping. Using a pixel-relative topological map representation, we propose \textbf{\emph{pixel-level} loop closures} that \emph{directly} enable relative-geometry grounded pixel-level planning of shorter paths. We establish these closures directly in pixel space, inserting them as zero-cost edges between geometrically grounded 3D pixel nodes. This seamlessly stitches overlapping topological regions into a unified spatial manifold, defined in a local relative 3D space, without requiring globally consistent reconstruction or pose graph optimization, as illustrated in Fig.~\ref{fig:teaser}.

This is in stark contrast with the typical role of loop closures in existing literature, where \textit{a)} in metric SLAM pipelines they typically account for drift in globally-referenced geometry, and \textit{b)} in image-topological pipelines they provide shortcuts but without any form of geometric leverage. As our topological shortening is densely grounded in pixel space, it yields dense pixel-level costmaps that encode accurate path planning costs. These costmaps are then used to condition a learnt-controller, providing the continuous geometric signals necessary for precise control prediction, an affordance that sparse image-relative baselines fundamentally lack~\cite{shah2023gnm, savinov2018semi, kwon2021visual, meng2020scaling, hahn2021no}. Through extensive simulated experiments, we demonstrate that \ours achieves an absolute improvement of over 35\% in both Success Rate and Success weighted by Path Length (SPL) compared to trajectory-bound baselines. Crucially, these simulation gains translate directly to physical hardware, with real-world mobile robot deployments demonstrating the practical efficacy of our approach.

In summary, this paper makes the following key contributions: \textit{i)} We introduce \textbf{\emph{pixel-level} loop closures} within a 3D relative mapping framework, seamlessly stitching disjoint topologies into a unified spatial manifold without global metric reconstruction.
\textit{ii)} We present a first-of-its-kind analysis of ground truth/simulator based loop detection and closures for the navigation task across image-, object- and pixel-level topological map representations.

\begin{figure*}
    \centering
    \includegraphics[width=0.9\linewidth]{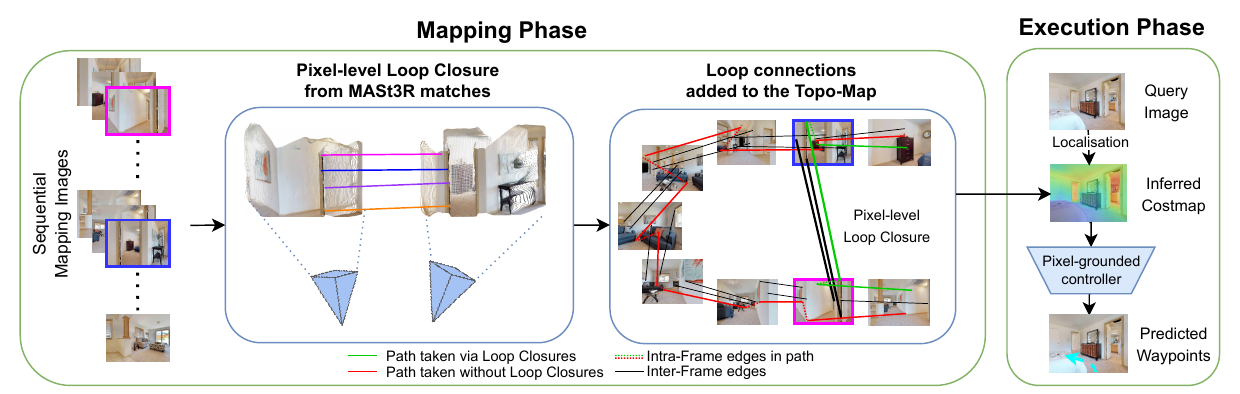}
    \caption{\textbf{Overview of the \ours navigation pipeline.}
\textbf{Left:} Offline mapping. Sequential images are processed with MASt3R to construct a topological graph with pixel-level loop closures.
\textbf{Right:} Online execution. The query image is localized and matched against the graph to generate a costmap that conditions the controller for waypoint prediction. \textit{Figure best viewed when zoomed in.}}
    \label{fig:pipeline}
    \vspace{-10pt}
\end{figure*}

\section{RELATED WORK}
Visual navigation research spans various representations, but our focus lies on topological frameworks that relax the need for explicit global metric SLAM~\cite{murORB2, engel14eccv}. We position our work by examining how existing map-less paradigms operate based on their node representations—image, object, and pixel—how they attempt to scale to arbitrary start-to-goal (A$\rightarrow$B) navigation, and why our dense loop closures provide a fundamentally superior scaling mechanism.

\paragraph{Image-Relative Topological Methods}
To avoid the computational overhead of globally consistent mapping, recent navigation models have widely adopted purely image-relative topological representations~\cite{savinov2018semi}. Methods such as ViNG~\cite{shah2021ving} and General Navigation Model (GNM)~\cite{shah2023gnm} completely discard explicit metric poses and global 3D maps. Instead, they abstract the environment into a graph where nodes are discrete images and edges denote visual similarity or predicted temporal proximity~\cite{shah2022rapid}. To scale beyond simple teach-and-repeat trajectories~\cite{furgale2010visual, vsegvic2009mapping} and support arbitrary A$\rightarrow$B navigation, these map-less frameworks attempt to establish loop closures by adding discrete scalar edges between visually similar but temporally distant views~\cite{shah2022viking, shah2022lmnav}. However, because connectivity is defined purely at the image level, these closures lack explicit geometric grounding. Handing the controller an entire discrete image as a sub-goal creates a severe information bottleneck, as it lacks the fine-grained, continuous geometric context necessary to smoothly steer through novel junctions~\cite{hutchinson1996tutorial, mezouar2002path}.

\paragraph{Object-Relative Topological Methods}
An alternative line of work seeks to ground navigation in semantically meaningful entities~\cite{conceptgraphs, armeni20193d}. Methods like ObjectReact~\cite{garg2025objectreact} and RoboHop~\cite{garg2024robohop} construct topological maps where nodes correspond to distinct objects or semantic regions, rather than whole images. These formulations can naturally link the same object observed across different traversals, forming semantic loop closures that aid in global route planning~\cite{wu2019bayesian, du2020learning}. While object-relative reasoning provides more localized sub-goals than full images~\cite{cai2024bridging, podgorski2025tango}, it inherently abstracts away the underlying structural geometry. The resulting path planning costs remain coarse, discarding the fine-grained geometric gradients (e.g., free space around the object) required for precise, collision-free control prediction through complex intersections.

\paragraph{Pixel-Relative Navigation and Loop-Augmented Scaling}
Recent geometry-aware approaches, such as \mnav~\cite{anonymous2026mast3rnav}, introduce denser topologies grounded in pixel-level relative 3D geometry~\cite{leroy2024grounding, duisterhof2025mast3r}. These representations preserve local spatial structure without demanding global metric consistency, but prior formulations are strictly constrained to single-run, teach-and-repeat trajectories, lacking any mechanism to reconcile multiple runs or exploit spatial overlap. \textbf{\ours} bridges this critical gap. We argue that loop closures are a superior mechanism for scaling topological maps to A$\rightarrow$B navigation, provided they are executed with geometric fidelity. By introducing loop closures directly in pixel space, we establish dense zero-cost edges between geometrically grounded 3D pixel nodes across image pairs. Unlike image-level loop closures that act as sparse proxy links, or object-level loop closures that discard structural nuances, our pixel-level loop closures directly reshape the geometric planning costmap. This fundamental structural shift provides the continuous geometric signals necessary for precise control prediction, enabling the robust discovery and exploitation of physically meaningful shortcuts across large environments.
\vspace{-5pt}
\begin{figure*}
    \centering
    \includegraphics[width=0.9\linewidth]{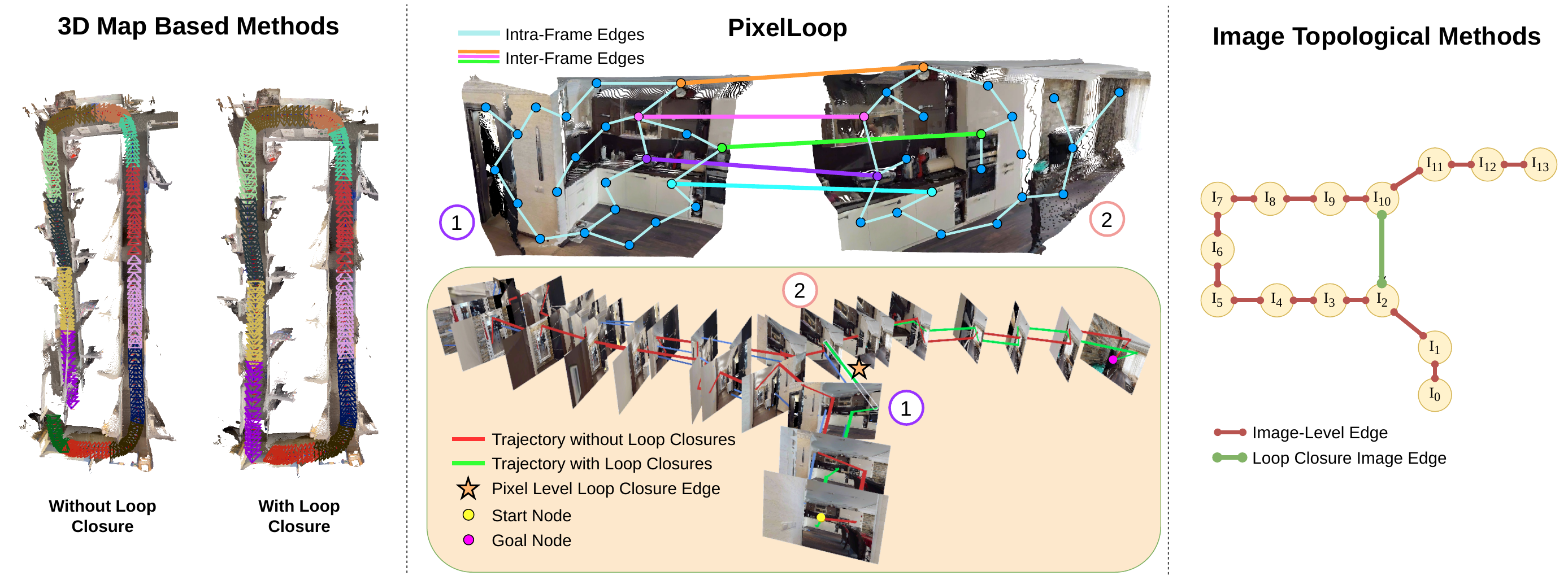}
    \caption{\textbf{Impact of loop closures across different map representations.} \textbf{Left:} 3D map-based methods use loop closures primarily to correct accumulated drift (visualized using VGGT-SLAM 2.0~\cite{maggio2025vggt-slam2}.) \textbf{Middle:} Our pixel-topological method establishes dense geometric shortcuts between loop closure candidates (\Circled{1} and \Circled{2}), enabling the agent to bypass the indirect sequential route (red) and take a shorter, direct path (green). \textbf{Right:} Image-topological methods establish a single discrete edge between image nodes (e.g., $I_2$ and $I_{10}$), lacking fine-grained geometric cues. \textit{(Absolute poses are used for visualization only; our method operates entirely in relative 3D space.)}}
    \label{fig:effect_of_loop_closure}
    \vspace{-15pt}
\end{figure*}
\vspace{-1pt}

\section{APPROACH}
\subsection{Background: Pixel Relative Navigation}
Recent work introduced \mnav, a visual navigation framework that leverages dense 3D correspondences to construct a relative-geometry grounded, pixel-level topological graph. Our complete pipeline (Fig.~\ref{fig:pipeline}) is fundamentally built upon this architecture. By explicitly relying on this pixel-level 3D map representation, \mnav preserves fine grained local geometry without demanding global metric consistency. 

\subsubsection{Pixel-Based Topological Mapping}
The foundation of this framework is the offline construction of a pixel-level topological graph from a sequence of reference images. During this offline mapping phase (Fig.~\ref{fig:pipeline} (Left)), incoming images are processed by MASt3R~\cite{leroy2024grounding}, a relative-geometry grounded image matching model. This model gives dense pixel correspondences and 3D pointmaps in a relative frame of reference between sequentially connected frames.

Specifically, for a given frame $I_t$, correspondences are computed within a temporal window $W$. These matches form the basis of the topological map $G=\{N,E\}$:
\begin{enumerate}
    \item \textbf{Nodes ($N$):} Each pixel successfully matched to another image is instantiated as a node in the graph.
    \item \textbf{Inter-frame Edges:} Pixels matched across different frames are connected by zero-cost edges, as they represent the same underlying 3D point observed from varying viewpoints. These edges are represented in black colour in Fig.~\ref{fig:pipeline} (Left).
    \item \textbf{Intra-frame Edges:} To allow path traversal through pixels within a single image's coordinate frame, edges are formed between distinct pixel nodes belonging to the same image. The cost assigned to these edges is the 3D Euclidean distance between their corresponding relative 3D coordinates as calculated using MASt3R pointmaps. To maintain computational tractability while preserving geometric structure, these connections are pruned to a Euclidean Minimum Spanning Tree (EMST), denoted as dotted lines in Fig.~\ref{fig:pipeline} (Left) and light-blue edges in Fig.~\ref{fig:effect_of_loop_closure} (Middle).
\end{enumerate}
The underlying pixel-level topological representation is analyzed in MASt3R-Nav~\cite[Table~IV]{anonymous2026mast3rnav}: while it is more resource intensive than its object-~\cite{garg2025objectreact} and image-based~\cite{shah2023gnm} counterparts, it remains tractable with a favourable accuracy-efficiency trade-off.

\subsubsection{Execution Phase}
During execution (Fig.~\ref{fig:pipeline} (Right)), the agent matches the current observation against the topological graph to generate a WayPixel~\cite{anonymous2026mast3rnav} costmap. A learned controller conditioned on this costmap then predicts a rollout of future 2D waypoints defined by their relative position and yaw in bird’s-eye-view (BEV) space.
\vspace{-4pt}

\subsection{PixelLoop}
In this section, we present our proposed pixel-level loop closure method, dubbed \textit{PixelLoop}, where we first describe \textit{loop detection}, which is composed of sequential descriptors based global retrieval and covisibility based verification using UniFlowMatch~\cite{zhang2025ufm} (UFM), and then present \textit{loop closure} and discuss in detail how pixel-level closures differ from the typical use and purpose of closures in globally-registered metric maps and image-level topological maps.

\subsubsection{Loop Detection}
While the standard mapping process connects sequential frames locally, it inherently lacks the ability to recognize when the camera revisits a previously mapped area. To address this, we identify temporally distant but geometrically overlapping views and establish loops between them.

\paragraph{Sequence Image Descriptors and Candidate Selection}
The first stage identifies broad structural similarities across the environment. Instead of relying on single-image descriptors, which are highly susceptible to local visual ambiguity, we extract robust sequence representations. We aggregate visual features over a sliding temporal window of 5 frames using a pretrained SeqVLAD~\cite{mereu2022learning} model (an extension of SeqNet~\cite{garg2021seqnet}) and compute a pairwise cosine similarity matrix between all frames. We define the set of initial loop closure candidates $\mathcal{C}_{cand}$ by applying a similarity threshold of $0.4$, selected empirically on a validation scene. To avoid matches between temporally adjacent frames that are already connected in the mapping window, we apply a temporal exclusion window of $\pm3$ frames around each query frame.

\paragraph{Dense Covisibility Check}
To verify the candidate pairs in $\mathcal{C}_{cand}$, we perform a bidirectional covisibility evaluation using UFM~\cite{zhang2025ufm}. Let $I_i$ denote the $i^{th}$ frame. First, UFM predicts a dense probability map assigning a visibility confidence to each pixel. A pixel in one image is classified as visible in the other if this confidence exceeds a fixed threshold of $0.9$. We then calculate the fraction of these visible pixels in both directions: from $I_i$ to $I_j$ and from $I_j$ to $I_i$. The final symmetric covisibility score, $\nu_{i,j}$, is defined as the minimum of these two directional fractions. A pair is successfully verified only if $\nu_{i,j} > 0.1$, ensuring sufficient 3D overlap between the two views.

This process yields a reliable set of loop closure candidates. Importantly, these detected loop image pairs are agnostic to the underlying mapping method. They establish a set of spatial connections that can be directly applied to any map representation, whether it is a global 3D metric map, an image-topological graph, or a pixel-topological graph.

\subsubsection{Loop Closure}
Once valid candidate pairs are identified, they must be integrated into the underlying map. While the detected loops themselves are representation-agnostic, the act of closing a loop alters different map representations in distinct ways. Consequently, these structural modifications dictate how the loops ultimately impact downstream path planning and navigation.

To contextualize our approach, we contrast how loop closures manifest across three distinct mapping paradigms (illustrated in Fig.~\ref{fig:effect_of_loop_closure}):

\paragraph{Global 3D Maps}
In traditional metric SLAM, loop closures primarily serve to reduce accumulated drift (Fig.~\ref{fig:effect_of_loop_closure} (Left)). They act as constraints in a pose graph optimization framework to reduce trajectory errors or jointly correct both global pose and map inconsistencies in the SLAM backend. Because the representation remains a rigid metric grid of occupied and free space, the impact on path planning is secondary, and loop closures mainly improve localization and mapping accuracy.

\paragraph{Image-Relative Topological Maps}
In purely topological frameworks, loop closures typically establish a single edge between discrete image nodes (Fig.~\ref{fig:effect_of_loop_closure} (Right)). Path planning relies on proxy metrics, such as predicted temporal distances, which depend on the model's implicit ability to infer spatial relationships and scene overlap. This reliance on implicit learning makes the system highly susceptible to perceptual aliasing and broader failures in the underlying image-matching task. Furthermore, providing the controller with an entire discrete image as a sub-goal creates a significant information bottleneck. The controller lacks fine-grained geometric cues indicating which specific regions of the image actually lead towards the target, often resulting in imprecise steering or navigation failures.

\paragraph{Pixel-Relative Topological Maps}
By leveraging MASt3R's dense correspondences, we establish numerous zero-cost inter-frame edges between matched pixels of $I_i$ and $I_j$. As visualized by the star edge in Fig.~\ref{fig:effect_of_loop_closure} (Middle), this effectively converts a standard image-level loop closure into a dense set of pixel-level loop closures. Rather than a single edge between two images, this injects fine-grained geometric shortcuts directly into the graph, enriching the pixel-level topology and altering the planning costmap.

\subsection{Path Planning and Control Prediction}
Exposing these dense geometric shortcuts to the controller effectively resolves the information bottleneck between the planner and the controller.
Because every matched pixel acts as a localized sub-goal, the system gains a larger bandwidth to tolerate matching errors or noisy sub-goals. Unlike image-level methods restricted to discrete image-pair servoing, dense pixel matching exposes granular, pixel-level path costs directly to the controller.

This structural shift directly improves navigation behavior by providing the planner with precise shortcuts. As illustrated in Fig.~\ref{fig:effect_of_loop_closure} (Middle), navigating from a start (yellow) to a goal (magenta) node using a strictly sequential map forces the agent along an indirect path (red trajectory). However, introducing pixel-level loop closure edges between the detected pair (\Circled{1} and \Circled{2}) creates direct pathways across the graph. This new connectivity propagates these shorter path distances throughout the graph, fundamentally altering the dense WayPixel costmaps. Consequently, the planner can bypass redundant sub-trajectories of the original teach-and-repeat route and compute a significantly shorter, direct path (green trajectory).

During the execution phase, the current query image is matched against this updated topological graph to produce a pixel-level query costmap. Because the dense costmap explicitly encodes these shortcuts, the controller can robustly predict waypoints along the shorter route, outperforming discrete image-level sub-goals. To ensure safe deployment around obstacles, these waypoint predictions are refined by a reactive Collision Avoidance via Repulsive Estimation (CARE)~\cite{kim2025care} module, paired with a history-based recovery mechanism that detects when the robot is stuck and reorients it towards open space. 

\section{EXPERIMENTS}

\subsection{Experimental Setup}

\textbf{Dataset.} We evaluate our approach on long ($\sim$100m on an average) indoor trajectories collected via offline teleoperation in 6 distinct HM3D~\cite{ramakrishnan2021habitatmatterport3ddatasethm3d} scenes. For each scene, we define three goal nodes and 3--5 start states per goal for evaluation, resulting in a total of 73 different start-to-goal navigation episodes. The trajectories are collected in an expansive manner, covering all navigable regions with diversity in camera orientations during revisits. As a result, each mapping sequence can be viewed as a composition of multiple interconnected teach-and-repeat sub-trajectories, as shown in Fig.~\ref{fig:teaser}.

From these trajectories, we construct topological graphs and corresponding WayPixel Costmaps as in Fig.~\ref{fig:pipeline} for planning and navigation evaluations.

\textbf{Navigation.} We evaluate start-to-goal navigation performance of \ours against three configurations: image-topological loop closures (GNM), object-topological loop closures (\objreact~\cite{garg2025objectreact}), and \mnav (i.e., without loop closures). At each timestep, the agent localizes against a local sub-map and selects the best matching reference frame for control. A trial is considered successful if the agent reaches within 0.5\,m of the goal node within 750 navigation steps. Collision avoidance using CARE is kept identical across all methods to ensure a fair comparison. Additionally, if the agent's displacement falls below 10\,cm over a sliding window of 15 consecutive steps, an estimated-depth-guided in-place recovery rotation of $45^\circ$ is triggered towards the direction with fewer obstacles in the top-down depth map.

\textbf{Evaluation Metrics.} We evaluate all methods using four standard embodied navigation metrics widely adopted in prior work~\cite{anderson2018evaluationmetrics, datta2020integratingegocentriclocalizationrealistic}. These include Success Rate (SR), Success weighted by Path Length (SPL, termed as SPL-A in our evaluations), SPL computed over successful episodes only (SPL-S), and Soft-SPL (SSPL), which assigns full credit to successful episodes and partial credit to failed episodes based on the agent’s final proximity to the goal.

\textbf{Ground Truth (GT)-Covisibility Loop Closures.} These are computed offline using ground-truth depth and camera poses. For each frame pair $(i,j)$ outside each other's mapping window (i.e., not already connected in the mapping graph), pixels in frame $i$ are backprojected to 3D and reprojected into frame $j$ using the known relative transform. A pair is declared a loop if the number of mutually consistent 3D points exceeds a predefined minimum (1000 points). Consistency is defined by the projected 3D location in $j$ agreeing with the ground-truth depth at that pixel within a small threshold (1 cm). We compare these against configurations without loop closures (None) and with our detected loop closures (SeqVLAD + UFM) in Tables~\ref{tab:mae_cost_study} and ~\ref{tab:combined_results}.

\begin{table*}[t]
\centering
\small
\setlength{\tabcolsep}{3.5pt}
\caption{Predicted cost MAE against two ground-truth references. 
Scores are computed over pixels within the minimum $k\%$ predicted cost 
and averaged across all mapping images. We observe lower MAE scores using loop closures detected using SeqVLAD + UFM than without.
\textbf{Bold: best overall. Underline: best non-GT} (lower is better).}
\label{tab:mae_cost_study}

\resizebox{\textwidth}{!}{
\begin{tabular}{lcccccccccccccccc}
\toprule

& \multicolumn{8}{c}{\textbf{Geodesic / Navigation Mesh-Based Costmaps}} 
& \multicolumn{8}{c}{\textbf{Ground Truth Pixel Correspondences}} \\

\cmidrule(lr){2-9} \cmidrule(lr){10-17}

\textbf{Loop Source}
& \multicolumn{4}{c}{Mean MAE $\downarrow$}
& \multicolumn{4}{c}{Median MAE $\downarrow$}
& \multicolumn{4}{c}{Mean MAE $\downarrow$}
& \multicolumn{4}{c}{Median MAE $\downarrow$} \\

\cmidrule(lr){2-5} \cmidrule(lr){6-9}
\cmidrule(lr){10-13} \cmidrule(lr){14-17}

& 5\% & 15\% & 30\% & 50\%
& 5\% & 15\% & 30\% & 50\%
& 5\% & 15\% & 30\% & 50\%
& 5\% & 15\% & 30\% & 50\% \\

\midrule

\rowcolor{gray!15}
GT-Covisibility
& 0.088 & \textbf{0.226} & \textbf{0.361} & \textbf{0.408}
& \textbf{0.099} & \textbf{0.259} & \textbf{0.369} & 0.385
& \textbf{0.073} & \textbf{0.174} & \textbf{0.262} & \textbf{0.299}
& \textbf{0.081} & \textbf{0.174} & \textbf{0.246} & \textbf{0.269} \\

\midrule

None
& 0.090 & 0.239 & 0.387 & 0.441
& 0.100 & 0.279 & 0.415 & 0.427
& 0.078 & 0.192 & 0.296 & 0.337
& 0.088 & 0.200 & 0.285 & 0.309 \\

SeqVLAD + UFM
& \textbf{0.087}
& \underline{\textbf{0.226}}
& \underline{\textbf{0.361}}
& \underline{0.409}
& \underline{\textbf{0.099}}
& \underline{0.260}
& \underline{0.370}
& \textbf{0.382}
& \underline{\textbf{0.073}}
& \underline{0.177}
& \underline{0.270}
& \underline{0.308}
& \underline{0.082}
& \underline{0.178}
& \underline{0.256}
& \underline{0.279} \\

\bottomrule
\end{tabular}
}
\vspace{-5pt}
\end{table*}
\vspace{-3pt}
\subsection{Evaluating Planning Costmaps}

We evaluate the cost-quality of our planning costmaps by comparing them against those produced without pixel-level loop closures. Specifically, we compute the Mean Absolute Error (MAE) between binary masks formed by selecting pixels corresponding to the minimum $k\%$ ($k \in \{5, 15, 30, 50\}$) of the predicted costmap and the ground-truth costmap. We use two forms of ground truth, as explained below:

\paragraph{Geodesic- / Navigation Mesh-Based Costmaps}  
 Using Habitat-Sim's~\cite{savva2019habitatplatformembodiedai} navigation mesh, we assign each pixel a cost equal to the geodesic distance between its closest navigable 3D point and the goal node.
    Concretely, each pixel is backprojected to its 3D world position using ground-truth depth and snapped to the nearest navigable point on the scene's navigation mesh.
    The geodesic distance from that snapped 3D point to the goal is then queried using Habitat-Sim's shortest-path solver and stored as the pixel's cost. These costs approximate true geodesic distances to the goal at each spatial location. MAE computed against this reference quantifies how closely the predicted costmaps approximate true geodesic distances.

\paragraph{Ground Truth Pixel Correspondences}  
    We generate ground-truth pixel correspondences by registering all pixels in global 3D space and subsampling matches (common 3D points in an image-pair) within an image using farthest-point sampling, while maintaining the same density as \master.
    These matches and ground-truth depth are then provided to our topological map construction pipeline, producing ground truth WayPixel costmaps, which are now unaffected by matching or depth estimation errors. MAE against this reference highlights path planning errors under perfect image matching but without the assumption of a globally-registered 3D map. 

Table~\ref{tab:mae_cost_study} reports the mean and median MAE scores aggregated across all mapping images and scenes. We compare costmaps generated without loop closures, with GT-Covisibility loop closures, and with loop closures detected using our pipeline (SeqVLAD + UFM).
We observe consistent reductions in MAE with the inclusion of loop closures across both the ground-truth references, indicating that pixel-level loop closures produce costmaps that more closely approximate ideal planning costs. Furthermore, the performance gap between loop-closure and no-loop configurations increases with larger values of $k$, suggesting that loop closures improve the global structure of the cost distribution. Notably, costmaps generated with loop pairs detected using SeqVLAD + UFM closely match those obtained using GT-Covisibility loop closures.

\begin{table}
\centering
\small
\setlength{\tabcolsep}{6pt}
\caption{Navigation performance across 73 start-to-goal tasks. 
SPL-S denotes SPL over successful episodes only, 
SPL-A denotes SPL over all episodes. We clearly observe higher metrics for \ours over other benchmarks with loops generated using SeqVLAD + UFM. Moreover, we observe a higher relative gain in performance on using pixel-level loop closures (as opposed to without), over image-level or object-level loop closures. \textbf{Bold: best overall. Underline: best non-GT per category} (higher is better).}
\label{tab:combined_results}

\begin{tabular}{lcccc}
\toprule
\textbf{Loop Source} & \textbf{SR} $\uparrow$ & \textbf{SPL-S} $\uparrow$ & \textbf{SPL-A} $\uparrow$ & \textbf{SSPL} $\uparrow$ \\
\midrule

\multicolumn{5}{c}{\textit{\ours (Ours)}} \\
\midrule

\rowcolor{gray!15}
GT-Covisibility & 54.79 & \textbf{91.70} & 50.25 & 58.39 \\
None & 35.62 & 80.45 & 28.65 & 40.70 \\
SeqVLAD + UFM & \underline{\textbf{68.49}} & \underline{91.15} & \underline{\textbf{62.43}} & \underline{\textbf{73.97}} \\

\midrule

\multicolumn{5}{c}{\textit{GNM~\cite{shah2023gnm} (HM3D)}} \\
\midrule

\rowcolor{gray!15}
GT-Covisibility & 24.66 & 70.50 & 17.38 & 18.40 \\
None & 20.55 & 64.80 & 13.32 & 15.95 \\
SeqVLAD + UFM & \underline{27.40} & \underline{76.97} & \underline{21.09} & \underline{22.44} \\

\midrule
\multicolumn{5}{c}{\textit{GNM-Adapted (HM3D)}} \\
\midrule

\rowcolor{gray!15}
GT-Covisibility & 32.88 & 83.56 & 27.47 & 28.01 \\
None & 17.81 & 75.28 & 13.41 & 16.95 \\
SeqVLAD + UFM & \underline{32.88} & \underline{83.97} & \underline{27.61} & \underline{28.56} \\

\midrule
\multicolumn{5}{c}{\textit{\objreact~\cite{garg2025objectreact}}} \\
\midrule

\rowcolor{gray!15}
GT-Covisibility & 63.01 & 65.21 & 41.09 & 65.91 \\
None & 58.90 & 63.21 & 37.23 & 38.77 \\
SeqVLAD + UFM & \underline{67.12} & \underline{69.37} & \underline{46.82} & \underline{68.35} \\

\bottomrule
\end{tabular}
\vspace{-15pt}
\end{table}
\vspace{-3pt}
\subsection{Navigation Comparisons}

We benchmark \ours's loop closures against image-level and object-level loop closure strategies under identical experimental settings. Specifically, we evaluate start-to-goal navigation performance by benchmarking \ours against GNM~\cite{shah2023gnm} (image-topological) and \objreact~\cite{garg2025objectreact} (object-topological), both with and without loop closures. This evaluation uses \master matches and estimated depth for \ours, while employing the reported state-of-the-art configurations for the other benchmarks.

We evaluate two GNM variants each using an HM3D-pretrained controller checkpoint provided by~\cite{garg2025objectreact}. a) \textit{GNM}, the publicly-released version, selects sub-goals directly at runtime based on its predicted temporal distances, without any graph-level path planning. b) \textit{GNM-Adapted}, which is our modification to GNM, augments this controller with a sequential image sub-graph akin to teach-and-repeat, achieved through Dijkstra-based shortest path computed over image-level edges weighted by GNM-predicted temporal distances; during execution, candidate sub-goals are restricted to a sliding window along this teach-and-repeat sub-graph, while localization continues to rely on GNM-predicted distances.
For the \objreact baseline, we use their state-of-the-art reported configuration. 

Our method outperforms image-level loop closures (GNM and GNM-Adapted) and object-level loop closures (\objreact) across all evaluation metrics presented in Table~\ref{tab:combined_results}. Incorporating pixel-level loop closures yields substantial improvements over configurations without loop closures, including a $\sim$34 point absolute increase in SPL-A (28.65 $\rightarrow$ 62.43). This gain is significantly larger than the improvement obtained by adding image-level loop closures to GNM-Adapted itself (14.20 points; 13.41 $\rightarrow$ 27.61). Similar comparisons can be seen against \objreact.

While \objreact achieves a competitive success rate (67.12 vs. 68.49), its SPL-A remains substantially lower (46.82 vs. 62.43), indicating that object-level loop closures enable recovery from incorrect trajectories but lack precise metric grounding in relative 3D geometry, often resulting in longer paths. This indicates that the performance gains stem not merely from introducing loop closures, but from the relative-geometry grounding enabled by pixel-level correspondences.

\begin{figure}
    \centering
    \includegraphics[width=\linewidth]{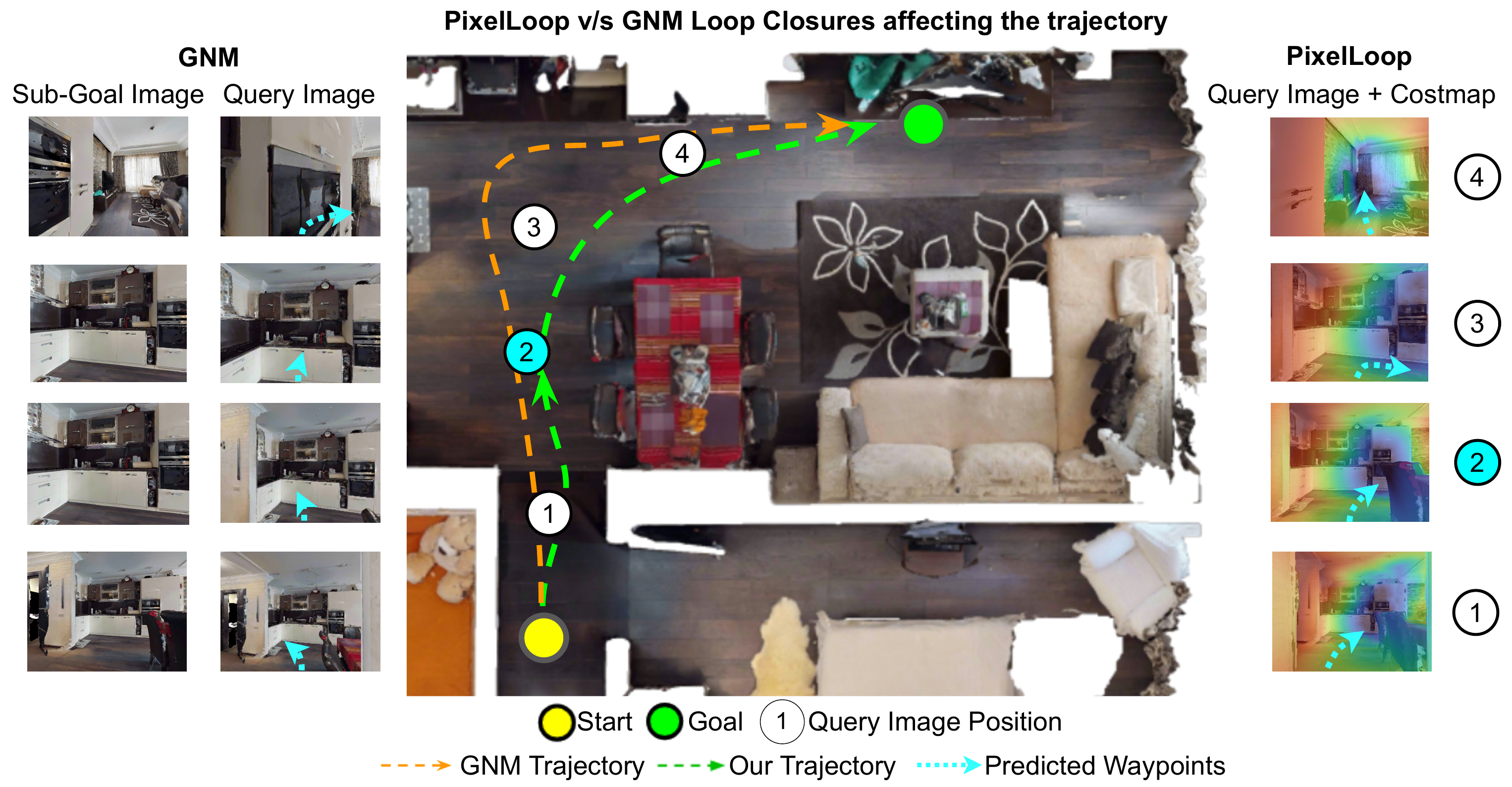}
    \caption{\textbf{GNM v/s \ours.} Both methods detect the same loop at point \Circled{2}. GNM treats the matched image as a holistic sub-goal, resulting in imprecise steering and trajectory drift. \ours produces smoother and more direct trajectories by leveraging relative-geometry grounded control signals.}
    \label{fig:pixel_vs_gnm}
    \vspace{-10pt}
\end{figure}

Fig.~\ref{fig:pixel_vs_gnm} provides a qualitative comparison between image-level loop closures (GNM) and our pixel-level loop closures during start-to-goal navigation. Although both methods detect the same loop at point \Circled{2}, their downstream control behavior differs significantly. GNM treats the matched image as a holistic sub-goal and predicts motion towards the entire view, without reasoning about which regions within the image actually lead towards the goal. Consequently, control signals are weakly specified, leading to imprecise steering and observable trajectory drift.

In contrast, \ours constructs dense pixel correspondences between the current view and the loop image. While not grounded in a global reference frame or metric map, these correspondences encode relative 3D geometric relationships between matched scene points. This induces a locally consistent relative-geometry representation, allowing the controller to infer waypoint targets that are spatially anchored towards the goal within the current view rather than directed towards the sub-goal image as a whole.

Importantly, these relative geometric relations propagate not only across loop closures but also along the toplogically connected map images, forming a cascading chain toward the goal node. This provides a structured control signal that consistently points toward the goal across viewpoints, a property unavailable to image-level loop closures, which lack pixel-level geometric grounding. As a result, \ours produces smoother and more direct trajectories, demonstrating that loop detection alone is insufficient unless translated into relative-geometry grounded control signals.

Performance using our loop detection pipeline surpasses even the GT-Covisibility loop closure configuration by a $\sim$24\% relative improvement in SPL-A. This improvement stems from pruning loop pairs with limited covisibility, which often share only small overlapping regions while the remaining image content contains visually similar yet geometrically inconsistent structures (e.g., doors on opposite sides of a wall), leading to incorrect matches.

\subsection{Real-World Demonstration}

\vspace{-4mm}
\begin{table}[h]
\centering
\small
\caption{Real-world navigation performance across 11 start-to-goal runs 
in 3 large indoor environments using a P3DX mobile robot with a RealSense camera.}
\label{tab:real_world_results}

\begin{tabular}{lccc}
\toprule
\textbf{Loop Source} & \textbf{SR} $\uparrow$ & \textbf{SPL} $\uparrow$ & \textbf{SSPL} $\uparrow$ \\
\midrule
None & 36.4 & 33.12 & 58.19 \\
\ours & \textbf{81.8} & \textbf{78.82} & \textbf{87.15} \\
\bottomrule
\end{tabular}
\end{table}
\vspace{-4mm}

\begin{figure}
    \centering
    \includegraphics[width=0.85\linewidth]{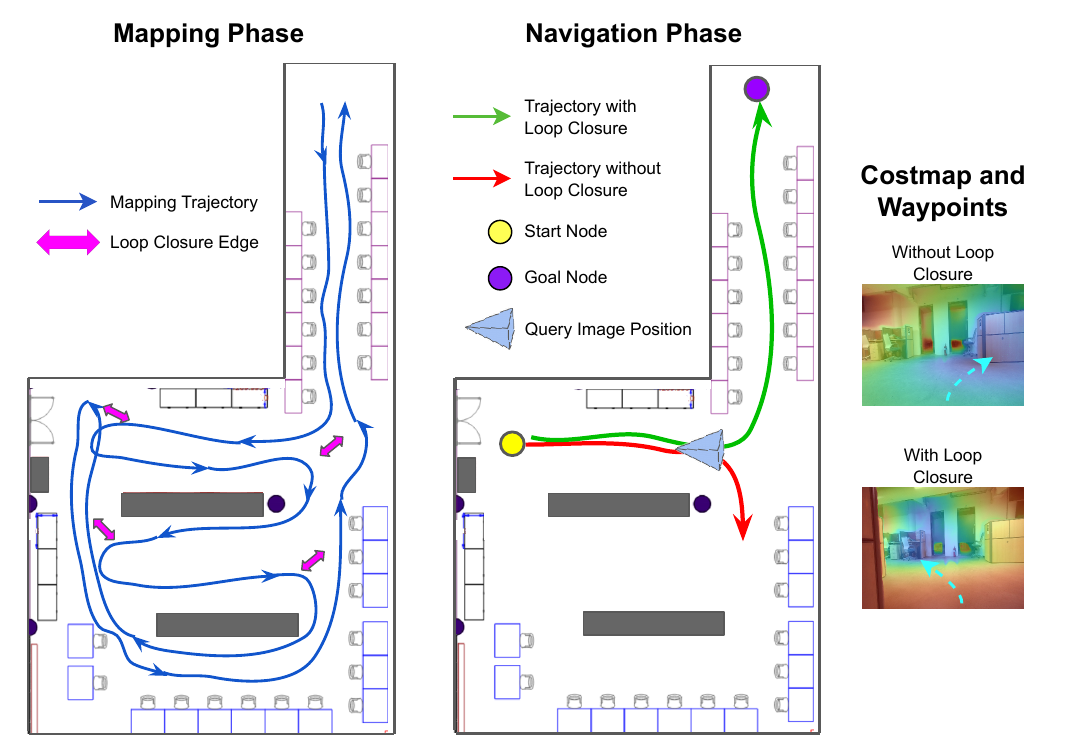}
    \caption{\textbf{Real World Demonstration.} We observe spatially accurate costmaps, controller waypoints and hence a shorter navigation trajectory in the presence of pixel-level loop closures as opposed to without.}
    \label{fig:rw_diagram}
    \vspace{-10pt}
\end{figure}

We further validate our approach in real-world indoor environments using a Pioneer 3-DX (P3DX) mobile robot equipped with an Intel RealSense camera. We evaluate \ours's navigation performance with and without pixel-level loop closures, measuring Success Rate (SR), SPL, and SSPL.

The results in Table~\ref{tab:real_world_results} and Fig.~\ref{fig:rw_diagram} demonstrate that the benefits of pixel-level loop closures extend beyond simulation and generalize to real-world deployments. PixelLoop runs at $\sim$4.8 Hz during online navigation on a single NVIDIA RTX A4000 GPU, requiring $\sim$4.6 GB of VRAM. MASt3R-based depth estimation and correspondence prediction jointly take $\sim$173.0 ms, while online query-costmap filling takes $\sim$6.7 ms at $320\times240$ resolution, with the remaining latency arising from pipeline overhead. Global SeqVLAD retrieval and UFM-based loop verification are performed offline. For a graph constructed using 350 RGB frames over a $\sim$100 m mapping trajectory, goal-conditioned preprocessing takes $\sim$1 min per goal, including $\sim$25 s for loop-aware shortest-path computation.

\section{CONCLUSION}
In this work, we introduced \textbf{\ours}, demonstrating that the key to robust topological navigation lies not merely in detecting loop closures, but in executing them with dense geometric fidelity. By operating directly in pixel space, our framework overcomes the information bottlenecks that severely limit discrete image-relative planners. We showed that stitching disjoint trajectories via zero-cost pixel correspondences fundamentally alters the underlying planning geometry, allowing us to generate continuous, fine-grained costmaps. 

Our empirical evaluations confirm that this structural shift empowers a local neural controller to reliably discover and exploit novel physical shortcuts. By achieving state-of-the-art performance in arbitrary start-to-goal navigation without the overhead of global metric SLAM, we validate the superiority of dense topological scaling. Crucially, successful real-world mobile robot deployments highlight the robustness and practical viability of our approach.
\vspace{-1pt}




\addtolength{\textheight}{-10cm}   








\bibliographystyle{IEEEtran}
\bibliography{references}

\end{document}